%% file: author.tex
\documentclass{svproc}
\usepackage{url}
\usepackage{graphicx}

\usepackage{graphicx}
\usepackage{subcaption}
\usepackage{placeins}   
\usepackage[table]{xcolor}
\usepackage{amsmath}
\usepackage{wrapfig}
\usepackage{wrapfig}
\usepackage[numbers]{natbib}
\usepackage{multicol}
\usepackage{tabularx}
\usepackage{booktabs}

\usepackage[bookmarks=true]{hyperref}
\usepackage{cleveref}

\usepackage{amsmath}
\usepackage{xcolor}
\usepackage{algorithm}
\usepackage{tikz}
\usetikzlibrary{arrows.meta, positioning, fit, calc, backgrounds}
\usepackage{graphicx}

\usepackage{multirow}
\usepackage{algorithmic}
\usepackage{graphicx} 
\usepackage{subcaption}
\usepackage{booktabs}
\usepackage{array}
\usepackage{tikz}
\usetikzlibrary{arrows.meta, positioning}
\usepackage{amssymb}  
\usepackage{cleveref}

\begin{document}
\mainmatter              
\title{Task-Conditioned   Uncertainty Costmaps for Legged Locomotion}
\titlerunning{Task-Conditioned Uncertainty Costmaps for
Legged Locomotion}  
%
\author{Kartikeya Singh\inst{1} \and Christo Aluckal\inst{1},
Romeo Orsolino\inst{2} \and Karthik Dantu\inst{1}}
\authorrunning{Singh et al.} 
%
\author{
Kartikeya Singh\inst{1}\thanks{Corresponding author.}
\and
Christo Aluckal\inst{1}
\and
Romeo Orsolino\inst{2}
\and
Karthik Dantu\inst{1}
}

\tocauthor{Kartikeya Singh, Christo Aluckal, Romeo Orsolino, and Karthik Dantu}

\institute{
DRONES Lab, University at Buffalo, NY, USA\\
\email{
ksingh35@buffalo.edu,
christoa@buffalo.edu,
kdantu@buffalo.edu
}
\and
Ocado Technology, UK\\
\email{orso.romeo@gmail.com}
}
\maketitle              

 \begin{abstract}

Legged robots maintain dynamic feasibility through multi-contact interactions with terrain. Learned foothold prediction can provide feasibility-aware costs for motion planning and path selection, but accurately predicting future contacts from perceptual inputs such as height scans remains challenging on highly unstructured terrain, even with a repetitive gait cycle. In this work, we show that modeling epistemic uncertainty in predicted footholds, conditioned on terrain observations and commanded motion, distinguishes in-distribution from out-of-distribution operating regimes in simulation and real-world settings. This allows a single learned model, trained on limited data distributions, to express uncertainty caused by missing training coverage. We use this learned uncertainty to detect OOD regions and incorporate them into a unified costmap-generation framework for uncertainty-aware path planning. 
Using these uncertainty-aware costmaps, we evaluate feasibility error across in-distribution and OOD terrains in simulation and real-world settings. The results show improved OOD detection, up to a \textbf{37\%} reduction in simulation feasibility error, and more reliable planning behavior than geometry-only baselines. 

\end{abstract}

\input{sections/intro}

\input{sections/related_work}

\input{sections/method}

\input{sections/experiments}

\input{sections/conclusion}

\begingroup
\let\clearpage\relax
\let\newpage\relax

\bibliographystyle{splncs04}
\bibliography{references}
\vspace{-4cm}
\endgroup

\end{document}

%% file: sections/intro.tex
\section{Introduction}

Legged robots are increasingly deployed in unstructured environments for tasks such as payload delivery~\cite{figliozzi2020autonomous}, search and rescue~\cite{bellicoso2018advances}, and environmental inspection~\cite{kolvenbach2019haptic}. However, autonomy stacks designed for wheeled robots typically do not reason about whether a planned trajectory will induce unstable or infeasible contact configurations for a legged robot. Learning-based planners face a related limitation: training data rarely covers the range of terrain geometry and commanded motion encountered during deployment.

A common heuristic is to treat terrain roughness as a proxy for locomotion risk~\cite{wermelinger2016navigation, wellhausen2021rough}. Roughness-based costs are useful, but they do not capture the task-dependent nature of legged locomotion. The same terrain patch may be feasible under slow, conservative motion and infeasible under faster or more dynamic commanded motion. Thus, geometric roughness alone cannot distinguish terrain that is intrinsically hard from terrain that becomes risky only under a particular locomotion command.

In this work, we address this gap by learning task-conditioned epistemic uncertainty~\cite{farid2022task} for future foothold prediction and using that uncertainty as a planning cost. Our model predicts future foothold positions from terrain observations and commanded base velocities while estimating epistemic uncertainty through stochastic inference. This uncertainty reflects limited support in the learned input--output mapping and is intended to increase when the model extrapolates to out-of-distribution terrain or motion regimes. To evaluate whether uncertainty-aware planning improves dynamic feasibility, we use feasibility margins derived from actuation-aware stability analysis~\cite{orsolino2020feasible, delprete2016fast}. In simulation and real-world experiments, task-conditioned uncertainty correlates with future foothold error and feasibility degradation, enabling more reliable path selection than geometry-only baselines.

Our main contributions are as follows:

\begin{itemize}
    \item We develop a foothold prediction model that conditions epistemic uncertainty on both height scans and commanded base velocities, allowing the same architecture to distinguish in-distribution (ID) and out-of-distribution (OOD) operating regimes.
    \item We train the uncertainty-aware foothold predictor in IsaacSim and deploy it on a quadruped robot in the real world without retraining.
    \item We use uncertainty-based foothold predictions to build ID/OOD costmaps and evaluate approximate future feasibility with legged-robot feasibility margins~\cite{orsolino2021rapid}.
    \item We integrate the uncertainty costmap with a custom MPPI planner and the NAV2 navigation stack, and evaluate planning performance in simulation and real-world experiments against geometry-only baselines.
\end{itemize}

%% file: sections/related_work.tex
\section{Related Work}

Safe off-road navigation with legged robots requires reasoning about locomotion stability while traversing complex terrain. Prior work has improved stability either through terrain-aware planning~\cite{elnoor2024pronav} or low-level optimization and stability analysis~\cite{orsolino2020feasible,orsolino2021rapid}. Model-based trajectory optimization can capture simplified robot dynamics~\cite{medeiros2020trajectory}, but often neglects practical feasibility constraints such as actuator torque and friction limits. Feasibility-based formulations~\cite{orsolino2020feasible} address these constraints and provide a principled framework for evaluating static or quasi-static stability, although accurately modeling dynamic locomotion in high-dimensional floating-base systems remains challenging.

\vspace{-2pt}

Prior work has also explored perception-driven foothold prediction and terrain-aware locomotion~\cite{belter2019single,zhang2018single}. Learning-based methods predict viable footholds directly from visual observations, while others evaluate or adapt footholds online based on terrain and robot dynamics~\cite{clemente2022foothold,ren2023hierarchical,tsounis2020deepgait}. However, these methods generally assume in-distribution operating conditions and do not explicitly model the epistemic uncertainty of future foothold predictions. In contrast, our approach estimates task-conditioned uncertainty in predicted footholds and uses it to identify unreliable regions for feasibility-aware planning under distribution shift.

Accurate feasibility estimation requires combining exteroceptive observations (e.g., RGB-D or LiDAR) with predictions of future robot dynamics, which is difficult because quadrupeds exhibit highly nonlinear behavior. Existing approaches estimate footholds using supervised labels or geometric/template-based methods~\cite{wellhausen2019should,wellhausen2021rough,asselmeier2024steppability}. Their reliability, however, degrades on non-coplanar terrain and under gait disturbances, making foothold prediction increasingly uncertain. We explicitly model this uncertainty and propagate it into traversability cost estimation for planning. The closest work is~\cite{queeney2025gram}, which models terrain-induced epistemic uncertainty to detect out-of-distribution (OOD) conditions and trigger a robust adaptation module. Our method extends this idea by jointly modeling uncertainty arising from both unseen terrain and commanded robot motion, rather than terrain observations alone.

%% file: sections/method.tex
\section{Uncertainty Approximation for Foothold Prediction on Uneven Terrain}

Given an unstructured scene, our objective is to approximate epistemic 
uncertainty in quadruped foothold prediction and provide it to a planner for 
uncertainty-aware planning. \Cref{Terrain-rep} describes the terrain 
representation, and \Cref{sec:training_epistemic} describes foothold estimation 
with uncertainty.

\subsection{Grid Representation for Terrain Characterization}
\label{Terrain-rep}
\begin{wrapfigure}{r}{0.5\linewidth}
     \vspace{-20pt}
     \centering
     \includegraphics[width=\linewidth]{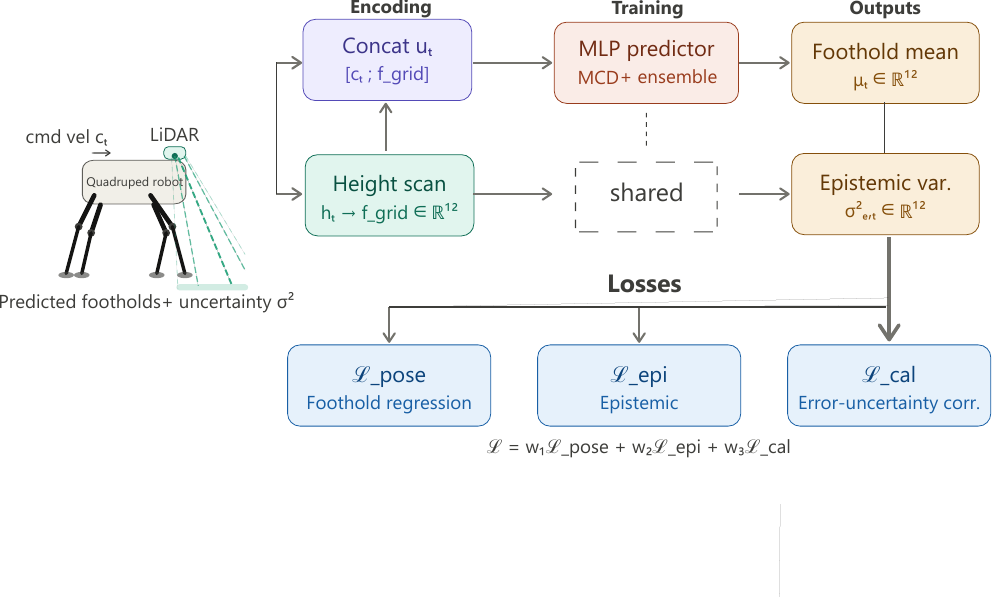}
     \vspace{-30pt}
     \caption{Overview of the training pipeline}
     \label{fig:overview}  
     \vspace{-25pt}

\end{wrapfigure}

We use a forward-facing LiDAR to perceive terrain in 3D, using only a 
 frontal slice as shown in~\Cref{fig:overview}. This region directly influences 
upcoming footholds and is encoded using a compact grid-based representation.

\noindent\textbf{Input Representation}: At each timestep, the terrain geometry is 
represented as a $6 \times 17$ grid of elevation values over a 
$0.6\,\text{m} \times 1.6\,\text{m}$ frontal region, sampled at $0.1\,\text{m}$ 
resolution, yielding 102 height measurements ($z_{i,j}$).

\noindent\textbf{Grid Pooling}: The frontal height grid is processed via 
non-overlapping average pooling into a compact terrain descriptor 
$\mathbf{f}_{\text{grid}} \in \mathbb{R}^{12}$, capturing coarse frontal terrain 
structure relevant for uncertainty estimation.

\subsection{Training with Epistemic Uncertainty}
\label{sec:training_epistemic}

We learn a supervised predictor mapping terrain observations at time $t$ to 
future foothold positions over horizon $H$. The supervision target is the 
stacked 3D positions of four feet in the robot base frame 
($\mathbf{y}_t \in \mathbb{R}^{12}$). Training rollouts are generated using a 
fixed pretrained locomotion policy~\cite{rudin2022learning} in IsaacSim. 
Epistemic uncertainty is estimated via stochastic inference combining ensemble 
methods with Monte Carlo (MC) dropout~\cite{chan2024estimating}.

\noindent\textbf{Input}: Two inputs are constructed: (i) a main input $\mathbf{x}_t$ 
for footprint mean prediction, and (ii) uncertainty-only input $\mathbf{u}_t$:
\begin{equation}
    \mathbf{u}_t =
    \Big[
    \underbrace{\mathbf{c}_t}_{\text{vel cmd (3)}}\,;
    \underbrace{\mathbf{f}_{\text{grid}}(\mathbf{h}_t)}_{\text{pooled height scan}}
    \Big],
    \label{eq:uncertainty_input}
\end{equation}
where $\mathbf{c}_t\in\mathbb{R}^3$ is the commanded base velocity and 
$\mathbf{f}_{\text{grid}}(\mathbf{h}_t)$ is the grid-pooled height scan defined in~\autoref{Terrain-rep}.

\noindent\textbf{Training:} A distributional mean--variance representation is 
consistent with the law-invariance principle (A6) in~\cite{majumdar2019should}, 
which states that risk depends only on the distribution of outcomes. Here, the 
model learns the predictive distribution and its epistemic uncertainty rather 
than defining a risk metric directly. The network predicts a footprint mean and epistemic 
 variance:
 \begin{equation}
    (\boldsymbol{\mu}_t,\;\boldsymbol{\sigma}^2_{e,t}) = 
    f_\theta(\mathbf{x}_t,\mathbf{u}_t),\quad
    \boldsymbol{\mu}_t\in\mathbb{R}^{12},\;
    \boldsymbol{\sigma}^2_{e,t}\in\mathbb{R}^{12}_{+}.
    \label{eq:model_outputs_epistemic}
\end{equation}
Using $K\!=\!3$ ensemble members and $M\!=\!20$ MC dropout samples (60 stochastic 
forward passes), the predictive mean and epistemic variance are:
\begin{equation}
    \bar{\boldsymbol{\mu}}_t = \frac{1}{KM}\sum_{k,m}\boldsymbol{\mu}^{(k,m)}_t,
    \quad
    \boldsymbol{\sigma}^2_{e,t} =
    \frac{1}{KM-1}\sum_{k,m}
    \!\left(\boldsymbol{\mu}^{(k,m)}_t - \bar{\boldsymbol{\mu}}_t\right)^{\!\odot 2},
    \label{eq:mc_ens}
\end{equation}
where $\odot$ denotes elementwise multiplication and $\boldsymbol{\sigma}^2_{e,t}$ 
is clamped to a safe numeric range during training.

\noindent\textbf{Loss}: We optimize three terms:
\begin{align}
    \mathcal{L} &= w_{\text{pose}}\mathcal{L}_{\text{pose}}
               + w_{\text{epi}}\mathcal{L}_{\text{epi}}
               + w_{\text{cal}}\mathcal{L}_{\text{cal}},
    \label{eq:total_epistemic_loss}\\
    \mathcal{L}_{\text{pose}} &= \tfrac{1}{12}
    \|\bar{\boldsymbol{\mu}}_t-\mathbf{y}_t\|_2^2,
    \quad
    \mathcal{L}_{\text{epi}} =
    \tfrac{1}{12}\sum_{j=1}^{12}
    \Big[(y_{t,j}-\bar{\mu}_{t,j})^2 - \sigma^{2}_{e,t,j}\Big]_+,
    \label{eq:pose_epi_loss}
\end{align}
where $[z]_+=\max(0,z)$ penalizes overconfidence. For calibration, we compute 
per-sample mean position error 
$e_t=\frac{1}{4}\sum_{i=1}^{4}\|\bar{\boldsymbol{\mu}}_{t,i}-\mathbf{y}_{t,i}\|_2$ 
and scalar uncertainty summary 
$\bar{s}_t=\frac{1}{12}\sum_{j}\sigma^{2}_{e,t,j}$. 
Within each minibatch, errors are linearly mapped to a target uncertainty band 
$[s_{\min},s_{\max}]$:
\begin{equation}
    s_t^\star = s_{\min} + (s_{\max}-s_{\min})\cdot
    \frac{e_t-e_{\min}}{(e_{\max}-e_{\min})+\epsilon},
    \label{eq:target_uncertainty}
\end{equation}
and the calibration loss enforces both absolute alignment and positive rank correlation, with
$\mathcal{L}_{\mathrm{cal}}=\mathbb{E}\!\left[\left|\bar{s}_t-s_t^{\star}\right|\right]
+\lambda\left(1-\rho(\mathbf{e},\bar{\mathbf{s}})\right)$\label{eq:lcal},
where $\rho(\mathbf{e},\bar{\mathbf{s}})$ is the Pearson correlation over minibatch vectors $\mathbf{e}=\{e_t\}_{t=1}^{B}$ and 
$\bar{\mathbf{s}}=\{\bar{s}_t\}_{t=1}^{B}$, and $\lambda$ controls the 
correlation regularizer strength.

\noindent\textbf{Inference Output}: At each timestep, the network outputs a 12-D 
foothold mean and a 12-D epistemic variance vector.

\section{Uncertainty-Aware Planning}

This section describes the uncertainty costmap derived from the learned 
uncertainty in~\Cref{sec:training_epistemic} and its integration into two motion 
planning frameworks. In simulation, we use a custom MPPI planner that directly 
optimizes trajectory costs using learned uncertainty. For real-world deployment, 
we integrate the costmap into the navigation stack from~\cite{cheng2022dynamic} 
on the Unitree Go1.

\subsection{Uncertainty Costmap Generation}
\label{sec:uncertainty-feasibility-costmap}

Per-leg epistemic uncertainty $\bar{\sigma}^2_{e,i}$ is obtained by averaging 
the three per-coordinate variances from $\boldsymbol{\sigma}^2_{e,t}$ 
(from \cref{eq:mc_ens}): $\bar{\sigma}^2_{e,i} = \frac{1}{3}\sum_{j \in \text{leg}_i} \sigma^2_{e,t,j},
    \quad i = 1,\ldots,4,$\label{eq:per_leg_uncertainty}. This value is converted to a cost $c_i = \min(100,\, \alpha\, \bar{\sigma}^2_{e,i})$ where 
$\alpha$ is a tunable scaling factor (for better thresholding). A Gaussian blob 
centered at each predicted foot location $(x_i, y_i)$ encodes spatial uncertainty 
decay:
\begin{equation}
    C(x, y) = \max_i \left( c_i \exp\!\left(
    -\frac{\|(x,y) - (x_i,y_i)\|^2}{2\sigma^2_b}\right) \right),
\end{equation}
where $\sigma_b$ is derived from a defined blob radius.

\subsection{MPPI-Based Navigation Planner}
\label{sec:mppi_planner}

We employ an MPPI planner to generate high-level velocity commands for
the Go1 in rough terrain, operating in a receding-horizon fashion. At
each step, $K$ control sequences are sampled, rolled out via a
discrete-time unicycle model, scored by a terrain-aware cost, and
combined via a softmin-weighted update.

\noindent\textbf{Kinematic model:}
\begin{equation}
\mathbf{x}_{t+1} =
\begin{bmatrix}
x_t + v_t \cos\psi_t\,\Delta t \\
y_t + v_t \sin\psi_t\,\Delta t \\
\psi_t + \omega_t\Delta t
\end{bmatrix},\quad
\boldsymbol{\tau}_{t}^{(k)} = \bar{\boldsymbol{\tau}}_{t}
+ \boldsymbol{\epsilon}_{t}^{(k)},\;
\boldsymbol{\epsilon}_{t}^{(k)}\!\sim\!\mathcal{N}(0,\Sigma),
\end{equation}
where $\boldsymbol{\tau}_t^{(k)}$ denotes the $k$-th sampled control sequence 
and $\bar{\boldsymbol{\tau}}_t$ is the nominal control.

\noindent\textbf{Terrain-aware cost:}
\begin{equation}
J^{(k)}=\sum_{t=1}^{H}\!\Big(
\lambda_g\|\mathbf{p}_t-\mathbf{p}_{\rm goal}\|
+\lambda_{\rm obs}c_{\rm obs}(\mathbf{p}_t)
+\lambda_r c_{\rm rough}(\mathbf{p}_t)
+\lambda_u C(\mathbf{p}_t)
+\lambda_{\rm ctrl}\|\boldsymbol{\tau}_t^{(k)}\|^2
\Big).
\end{equation}
where $\lambda_g$ encourages goal progress and $\lambda_{\rm ctrl}$ penalizes 
control effort (both active in all configurations). Exactly one of the three 
middle terms is active per planning run: baseline~(i) uses only 
$\lambda_{\rm obs} c_{\rm obs}(\mathbf{p}_t)$, a fixed height-threshold obstacle 
cost; baseline~(ii) uses only $\lambda_r c_{\rm rough}(\mathbf{p}_t)$, a 
height-scan variance roughness cost; and the proposed method uses only 
$\lambda_u C(\mathbf{p}_t)$, the learned epistemic uncertainty costmap from 
\cref{eq:per_leg_uncertainty}. The remaining two weights are set to zero 
accordingly.

\noindent\textbf{Control update:}
\begin{equation}
w^{(k)}=\frac{\exp(-J^{(k)}/\beta)}{\sum_{i}\exp(-J^{(i)}/\beta)},
\qquad
\bar{\boldsymbol{\tau}}_t \leftarrow \sum_{k}w^{(k)}\boldsymbol{\tau}_t^{(k)},
\end{equation}
where $\beta$ is the temperature parameter. The first control is
executed, and the horizon shifts forward. The controller update naturally generates the rollouts with velocities outside the training distribution that we utilize in the ID/OOD detection during our real-world experiments.  

\section{Feasibility Evaluation Using Proposed Uncertainty}
\label{sec:feas_eval_nongrad}

To evaluate robot feasibility while traversing the planned path in both MPPI and 
NAV2 runs, we adapt stability margins from~\cite{orsolino2020feasible,orsolino2021rapid} 
as the metric. We compare \emph{predicted} vs.\ \emph{actual} footholds. At each 
timestep $t$, predicted footholds 
 $\hat{\mathbf{p}}_t \in \mathbb{R}^{4 \times 3}$ are transformed to world frame 
and compared against simulator contact positions $\mathbf{p}_t$. For real robot, we obtain actual footholds via forward kinematics. Margins are computed via the iterative projection (IP) method,
$m^{\mathrm{pred}}_t = \mathrm{IP}(\hat{\mathbf{p}}_t)$ and
$m^{\mathrm{actual}}_t = \mathrm{IP}(\mathbf{p}_t)$,
where $m>0$ indicates feasibility and $m<0$ indicates infeasibility. Margins are mapped to scalar costs
$c(m)=\left\{\begin{array}{ll}
\dfrac{1}{m+\epsilon}, & m>0,\\
|m|+1, & m\le0,
\end{array}\right.$
with $\epsilon=0.01$. The per-step feasibility error 
$e_t = |c(m^{\mathrm{pred}}_t) - c(m^{\mathrm{actual}}_t)|$ 
is averaged over each trajectory; lower values indicate better alignment with 
true robot stability.

%% file: sections/experiments.tex
\section{Ablation}
\label{sec:Ablation}

\begin{wrapfigure}{r}{0.5\linewidth}
    \vspace{-29pt}
    \includegraphics[width=\linewidth]{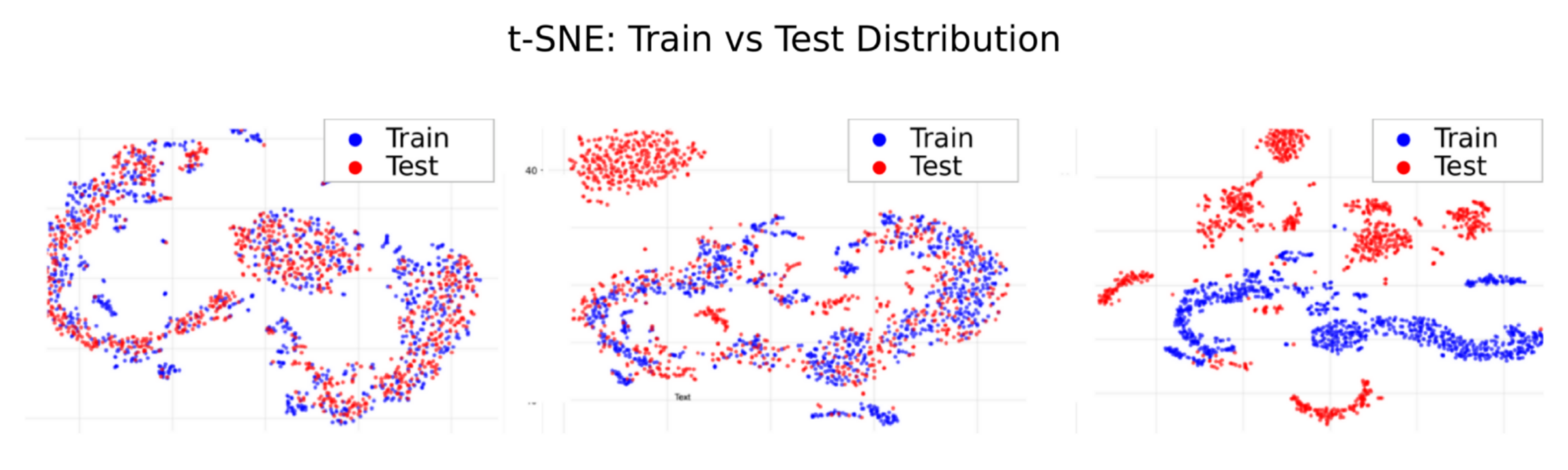}
    \caption{t-SNE of terrain input distribution shift: training (red) vs.\
    OOD test (blue). \textbf{Left:} ID features. \textbf{Middle:} OOD
    terrain, ID velocity. \textbf{Right:} OOD terrain, OOD velocities.}
    \label{fig:raw_distribution_shifts}
    \vspace{-22pt}
\end{wrapfigure}

In this section, we analyze how different input sources affect predicted uncertainty and
its correlation with foothold prediction error. \Cref{fig:raw_distribution_shifts} shows the t-SNE distribution shift
between training and three OOD conditions: same features as training
(left), OOD terrain with ID velocity (middle), and OOD terrain with
varying velocities (right). The rightmost plot shows the additional shift induced by OOD velocities, which this work intends to account for.

\begin{wrapfigure}{r}{0.5\linewidth}

\centering
\includegraphics[width=0.49\linewidth,
    trim=0.15cm 1.5cm 0.15cm 1.5cm,clip
]{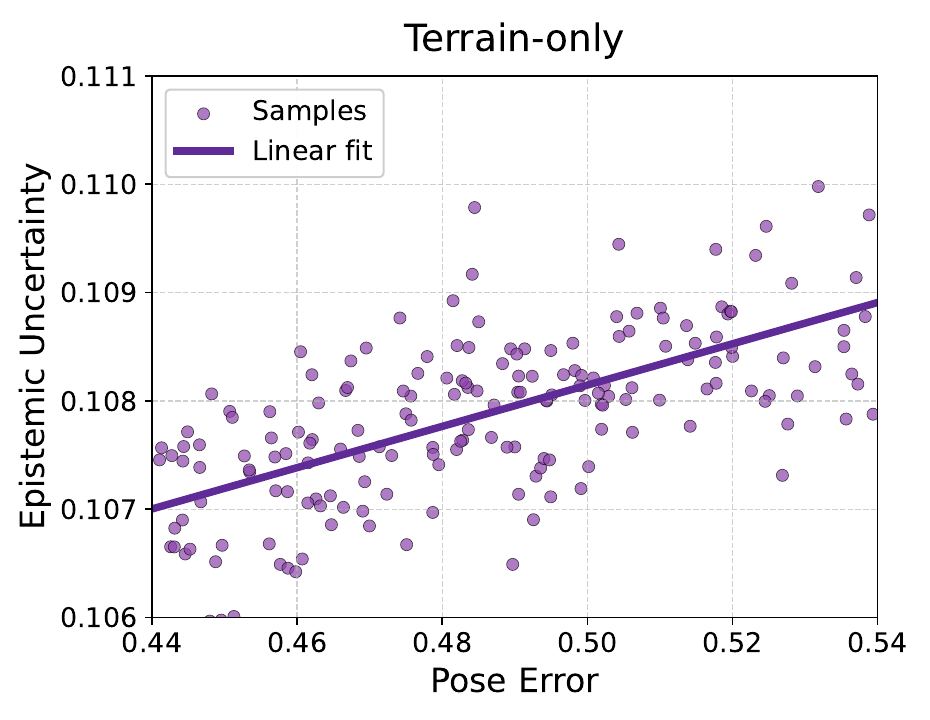}%
\hfill%
\includegraphics[width=0.49\linewidth,
    trim=1.0cm 1.5cm 0.1cm 1.5cm,clip
]{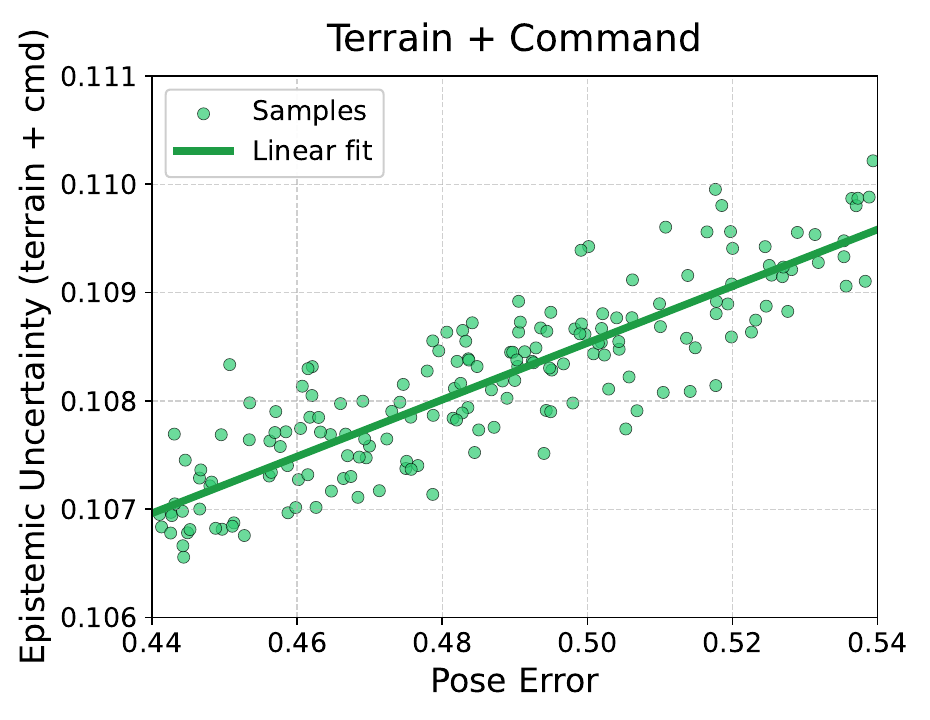}
\par\smallskip
\includegraphics[width=0.49\linewidth,
    trim=0.15cm 1.6cm 0.25cm 1.4cm,clip
]{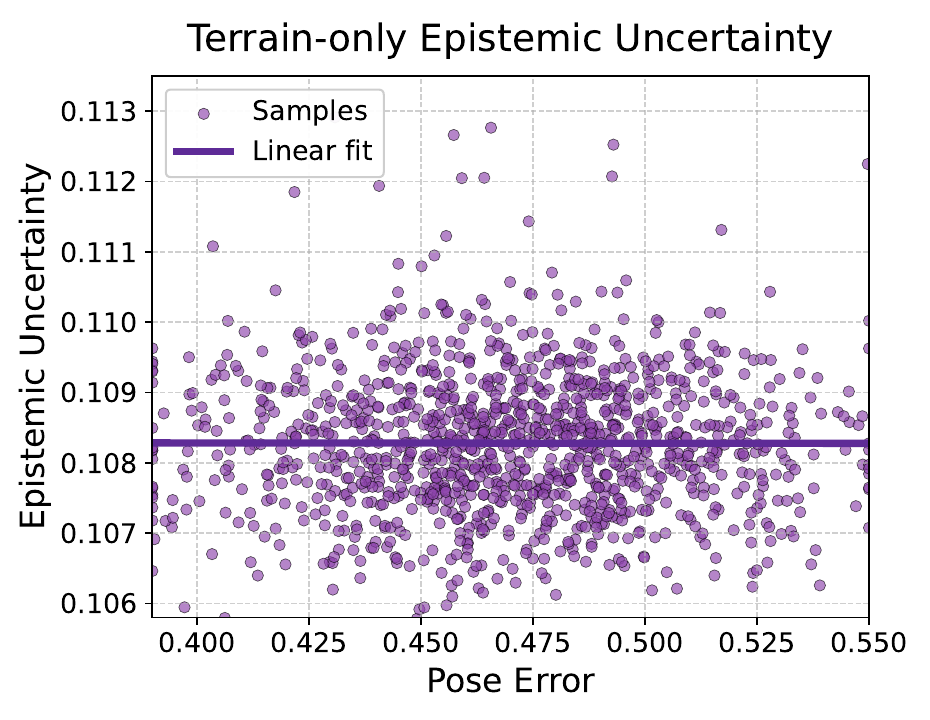}%
\hfill%
\includegraphics[width=0.49\linewidth,
    trim=1.0cm 1.6cm 0.35cm 1.6cm,clip
]{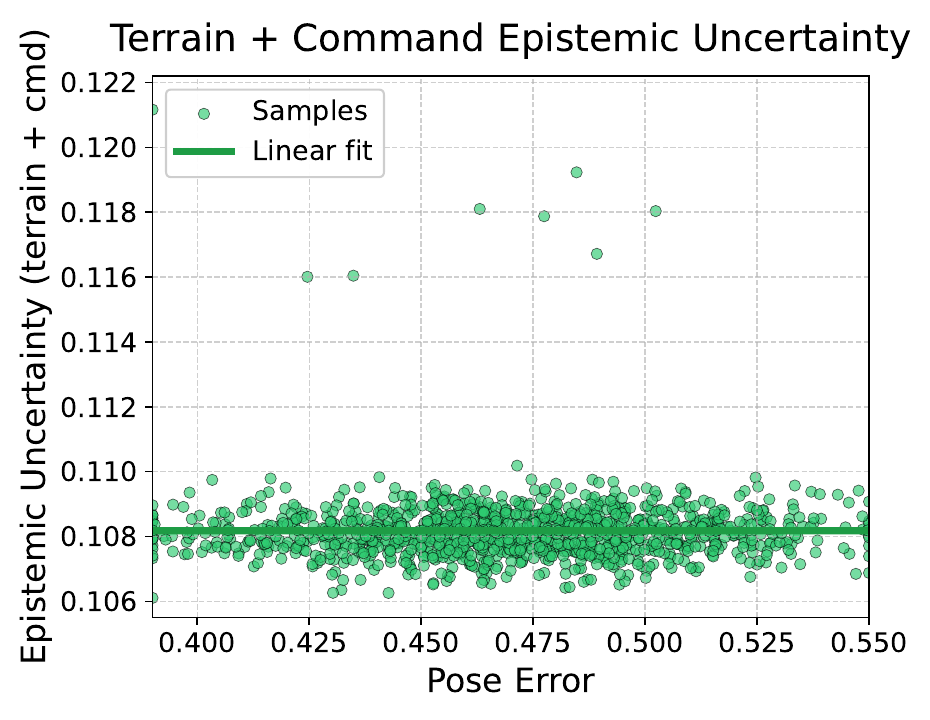}
\vspace{-9pt}
\caption{\small Uncertainty--error correlation: terrain-only (left)
vs.\ Proposed (right).}
\label{fig:horizon_corel_corr}
\vspace{-25pt}
\end{wrapfigure}

\Cref{fig:horizon_corel_corr}
shows the learned uncertainty-- foothold error correlation on OOD inputs: our
terrain+cmd formulation (right column) shows a steeper positive slope
and a tighter linear fit than the terrain-only baseline (left column),
confirming that conditioning on commanded velocity, along with the terrain roughness, improves calibration. A key distinction between our approach and heuristic OOD detectors based
on height-scan variance~\cite{wellhausen2021rough,wermelinger2016navigation}
is what uncertainty represents. Height-variance methods operate purely in
input space, which remains agnostic to the downstream task and insensitive to whether
geometric deviations actually degrade prediction accuracy. In contrast,
our epistemic uncertainty is task-conditioned and calibrated to prediction
error: trained only on flat terrain with a limited range of commanded velocities, the model yields low uncertainty
in-distribution and naturally elevated uncertainty under distribution
shift from unseen terrain or velocities.

\section{Experiments}
\label{sec:Path Planning Experiments}

\subsection{Experiment Setup}
\label{subsec:setup}

The IsaacSim-trained model is deployed to simulation and the real world
without retraining. Training uses the Unitree Go1 with a
height scanner on flat terrain at constant velocity. For real-world
deployment, a downward-facing LiDAR on the robot head replaces the
simulated scanner. Given point cloud
$\mathcal{P}=\{(x_k,y_k,z_k)\}$ in robot frame, we synthesize a
$6{\times}17$ height-scan grid over window
$x\!\in\![0,S_x],\;y\!\in\![-S_y/2,\,S_y/2]$, flattened row-major to
form the frontal height-scan vector. We evaluate the ID/OOD uncertainty across six terrain
configurations: three simulation terrains (wavy, stepped, spiked rough)
each repeated over three runs from different starting points, and three real-world platforms (5\,cm
raised, 10\,cm raised, and a variable-height ramp of 0--15\,cm). However, the planner terrain setup had mixed terrains and is shown in~\Cref{fig:planner} as a colored heightmap.

We train a single model in IsaacSim on flat terrain with constant
velocities (ID set). Conditions inducing distribution shift constitute
OOD; elevated uncertainty under OOD is enforced via
$\mathcal{L}_{\rm cal}$ (\Cref{eq:lcal}). \noindent\textbf{Baselines:}
\emph{(i)~Terrain-only uncertainty (Roughness)}: height-scan-variance OOD detection
treating high variance regions as lethal, independent of commanded
motion.
\emph{(ii)~Obstacle-only}: avoids obstacles via a fixed
height threshold, without uncertainty modeling.
``Baseline Uncertainty'' refers to~(i) unless stated otherwise.

\begin{figure}
\vspace{-1pt}
\centering
\includegraphics[width=\linewidth]{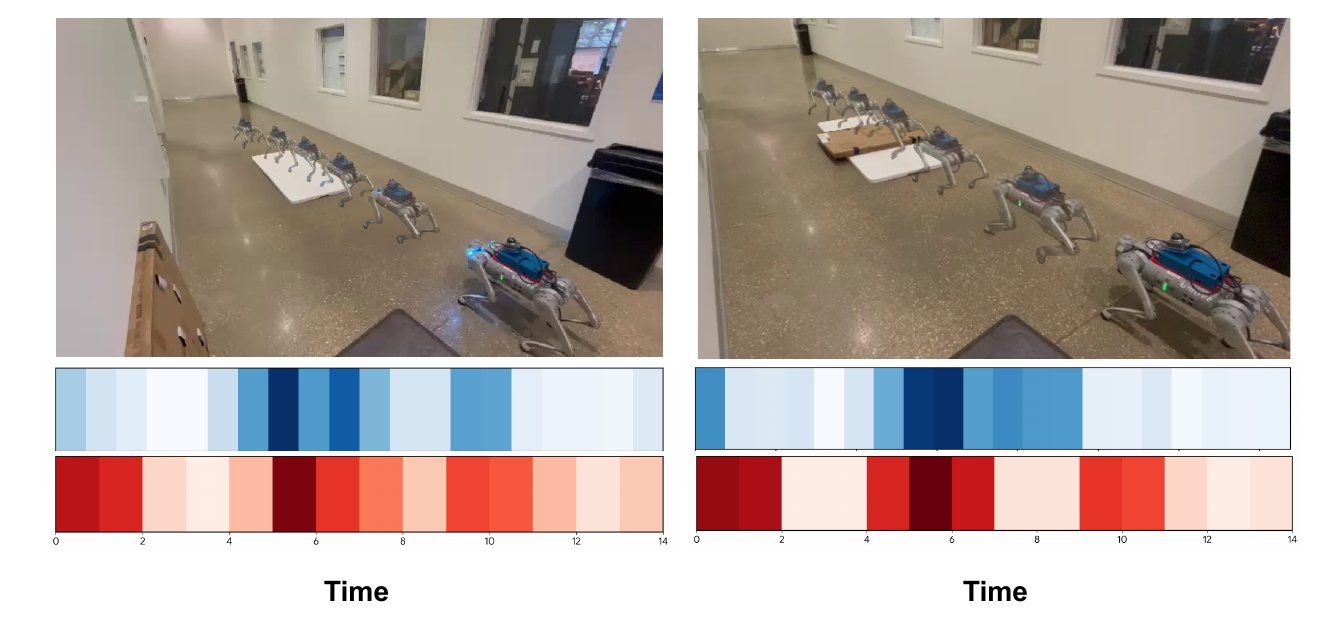}
\vspace{-21pt}
\caption{\small \textcolor{blue}{Terrain-only} vs 
\textcolor{red}{Proposed} uncertainty. Darker shade = uncertainty 
spike.}
\label{fig:redvsblue}
\vspace{-23pt}
\end{figure}

\subsection{ID–OOD Distinction}
\label{subsec:uncertainty_vs_variance}

We assess whether the proposed epistemic uncertainty reliably 
distinguishes ID from OOD terrain regions, and whether detected OOD 
regions correspond to genuinely harder footholds. At each timestep 
$t$, the proposed method computes the scalar epistemic summary $\bar{s}_t$ 
(from \cref{eq:mc_ens}) and the terrain-only baseline computes 
height-scan variance $\text{Var}(\mathbf{h}_t)$. Each method's base 
threshold is the mean of its signal over ID training trajectories:
$
    \theta_{\text{prop}} = \mathbb{E}_{\text{ID}}\!\left[\bar{s}_t\right],
    \qquad
    \theta_{\text{terrain}} = 
    \mathbb{E}_{\text{ID}}\!\left[\mathrm{Var}(\mathbf{h}_t)\right].
    \label{eq:ood_thresholds}
$
Contiguous runs exceeding the threshold are grouped into candidate OOD 
segments.

The top-$K$ segments by mean uncertainty are labeled OOD and the remaining segments ID, where $K$ equals the number of distinct terrain  transitions in the test environment. A well-calibrated signal yields lower error in ID and higher error in OOD regions producing a larger ID$\downarrow$/OOD$\uparrow$ 
gap. \Cref{fig:redvsblue} shows the two signals over a 5cm and 10cm planks traversal; the proposed uncertainty spikes at multiple transitions caused by terrain variation and varying velocities, while the terrain-only signal responds mainly near the raised platform.

\Cref{tab:feasibility-uncertainty-comparison} reports foothold 
prediction error within detected ID and OOD regions. Our method 
consistently achieves a larger ID$\downarrow$/OOD$\uparrow$ gap across 
all three simulation terrains (wavy, stepped, spiked), confirming that 
detected OOD regions correspond to genuinely harder footholds. Lower 
OOD error in baseline-detected regions suggests missed detections, with difficult footholds incorrectly classified as ID segments.
\vspace{-8pt}

\begin{table}[t]
\centering
\caption{Foothold error (cm) in ID/OOD regions across terrains with 3 different starting points (rows). Our method achieves a larger ID$\downarrow$/OOD$\uparrow$ gap.}
\label{tab:feasibility-uncertainty-comparison}
\scriptsize
\setlength{\tabcolsep}{2pt}
\renewcommand{\arraystretch}{1.03}

\begin{tabular}{c@{\hspace{1.5em}}c@{\hspace{1.5em}}c}
\multicolumn{1}{c}{\textbf{Simulation}} &
\multicolumn{1}{c}{|} &
\multicolumn{1}{c}{\textbf{Real-world}} \\
\end{tabular}

\vspace{2pt}

\begin{minipage}{0.47\columnwidth}
\centering
\begin{tabular}{c|cc|cc|cc}
\toprule
 & \multicolumn{2}{c|}{Wavy} 
 & \multicolumn{2}{c|}{Step}
 & \multicolumn{2}{c}{Spike} \\
\cmidrule(lr){2-3}\cmidrule(lr){4-5}\cmidrule(lr){6-7}
 & ID & OOD & ID & OOD & ID & OOD \\
\midrule
Base  & 18 & 20 & 18 & 20 & 18 & 20 \\
Ours  & \cellcolor{green!18}$\mathbf{17}$ & \cellcolor{green!18}$\mathbf{21}$ & \cellcolor{green!18}$\mathbf{17}$ & \cellcolor{green!18}$\mathbf{21}$ & \cellcolor{green!18}$\mathbf{17}$ & \cellcolor{green!18}$\mathbf{21}$ \\
\midrule
Base  & 16 & 18 & 16 & 19 & 17 & 19 \\
Ours  & 16 & \cellcolor{green!18}$\mathbf{19}$ & 16 & \cellcolor{green!18}$\mathbf{19}$ & 17 & \cellcolor{green!18}$\mathbf{19}$ \\
\midrule
Base  & 17 & 14 & 17 & 19 & 17 & 20 \\
Ours  & 17 & \cellcolor{green!18}$\mathbf{19}$ & 17 & \cellcolor{green!18}$\mathbf{20}$ & 17 & \cellcolor{green!18}$\mathbf{27}$ \\
\bottomrule
\end{tabular}
\end{minipage}
\hfill
\vrule width 0.6pt
\hfill
\begin{minipage}{0.47\columnwidth}
\centering
\begin{tabular}{c|cc|cc|cc}
\toprule
 & \multicolumn{2}{c|}{5cm} 
 & \multicolumn{2}{c|}{10cm}
 & \multicolumn{2}{c}{Ramp} \\
\cmidrule(lr){2-3}\cmidrule(lr){4-5}\cmidrule(lr){6-7}
 & ID & OOD & ID & OOD & ID & OOD \\
\midrule
Base  & \cellcolor{green!18}$\mathbf{15}$ & 27 & 12 & 09 & 13 & 06 \\
Ours  & 16 & \cellcolor{green!18}$\mathbf{34}$ & 12 & \cellcolor{green!18}$\mathbf{12}$ & 13 & \cellcolor{green!18}$\mathbf{13}$ \\
\midrule
Base  & 16 & 11 & 18 & 14 & 28 & 12 \\
Ours  & 16 & \cellcolor{green!18}$\mathbf{26}$ & 18 & \cellcolor{green!18}$\mathbf{16}$ & \cellcolor{green!18}$\mathbf{23}$ & \cellcolor{green!18}$\mathbf{15}$ \\
\midrule
Base  & 18 & 13 & 13 & 12 & \cellcolor{green!18}{$\mathbf{13}$} & 11 \\
Ours  & 18 & \cellcolor{green!18}{$\mathbf{16}$} & 13 & \cellcolor{green!18}{$\mathbf{18}$} & 28 & \cellcolor{green!18}{$\mathbf{21}$} \\
\bottomrule
\end{tabular}
\end{minipage}
\vspace{-0.5cm}
\end{table}

\subsection{Feasibility and Costmap Evaluation}
\label{subsec:feasibility_eval}

\begin{figure}
    \centering
    \vspace{-15pt}
    \includegraphics[
        width=\linewidth
    ]{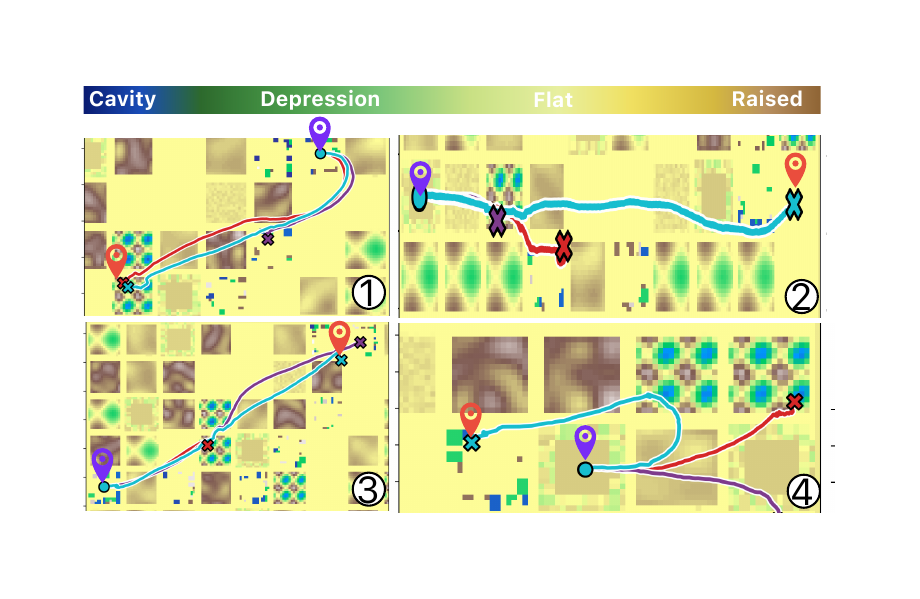}
    \vspace{-50pt}
    \caption{Progressive effect of uncertainty weighting on planning. From (1) to (4), increasing uncertainty cost shifts the planner from aggressive, risk-prone trajectories to conservative paths that avoid uncertain terrain.}
    \label{fig:planner}
    \vspace{-20pt}
\end{figure}

\begin{figure}[t]
\centering

\includegraphics[width=\linewidth]{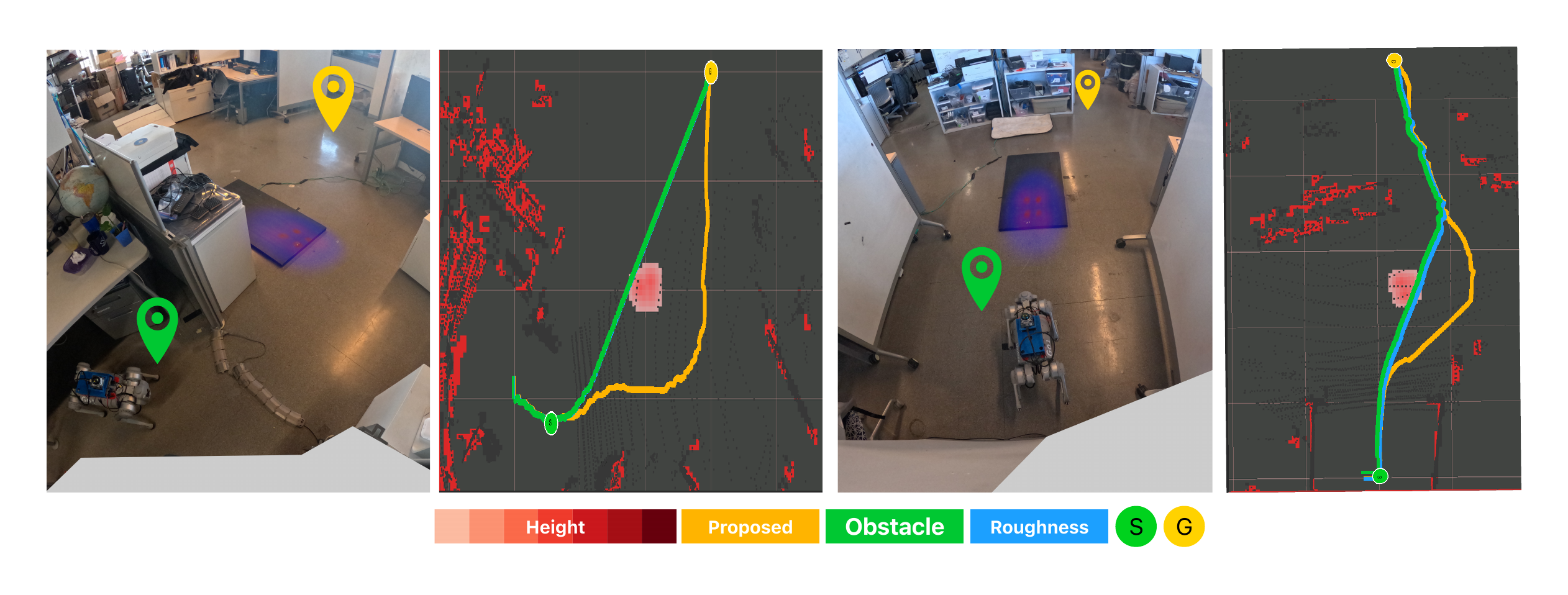}
\vspace{-20pt}
\caption{\small Planning results on the Unitree Go1 using NAV2. The purple patch denotes the OOD region induced by a 5cm board (velocity and terrain roughness).}
\label{fig:realworld_costmap}

\end{figure}

We evaluate the feasibility of the path traversed by the robot as explained in \Cref{sec:feas_eval_nongrad}. In this experiment, we compare our proposed path with two baselines: (i) obstacle-only, which uses a fixed height 
threshold, and (ii) roughness-only, which uses height-scan variance as 
a lethal cost. Evaluation is conducted in both simulation, using the 
mixed terrain shown in \Cref{fig:planner}, and in the real world on 
the Unitree Go1 via NAV2 in~\Cref{fig:realworld_costmap}.

We run 10 independent MPPI planning 
experiments per costmap formulation, each from a randomized start 
configuration on the mixed simulation terrain. Feasibility error 
$e_t = |c(m^{\mathrm{pred}}_t) - c(m^{\mathrm{actual}}_t)|$ is 
averaged over each trajectory (from \Cref{sec:feas_eval_nongrad}), 
and reported as mean $\pm$ std across timesteps. Lower feasibility 
error indicates better alignment between the planned path and true 
robot stability.\Cref{tab:feasibility_error_summary} reports feasibility error across
10 planning experiments on the mixed simulation terrain shown
in~\Cref{fig:planner}.

Each experiment corresponds to an independent
planning run from a randomized start configuration. Our uncertainty-based
costmap achieves the lowest mean feasibility error, with a 37\% lower mean error,
outperforming both the obstacle-only and roughness-only baselines.

\Cref{fig:planner} shows the progressive effect (from 1 to 4 in increasing order of uncertainty threshold) of uncertainty
weighting on planning paths
for all three methods across representative terrain configurations
containing flat regions, depressions, raised terrain, and pits. The
proposed cost (blue path) consistently routes through lower-uncertainty
regions, while the obstacle-only baseline (red path) takes a shorter but
riskier paths, and the roughness baseline (purple path) is overly
conservative on geometrically rough but dynamically feasible terrain.

\begin{wraptable}{l}{0.48\linewidth}
\vspace{-20pt}
\centering
\caption{Feasibility error (Mean\,$\pm$\,Std) for 10 MPPI runs.}
\label{tab:feasibility_error_summary}
\scriptsize
\setlength{\tabcolsep}{2pt}
\renewcommand{\arraystretch}{1.05}
\begin{tabular}{cccc}
\toprule
Exp & Obs & Rough & Ours \\
\midrule
1  & $16.2\pm29.1$   & $21.7\pm27.7$   & \cellcolor{green!18}$\mathbf{15.2\pm22.3}$ \\
2  & $17.0\pm31.7$   & $17.3\pm31.2$   & \cellcolor{green!18}$\mathbf{15.2\pm22.3}$ \\
3  & $15.9\pm28.1$   & $21.0\pm29.9$   & \cellcolor{green!18}$\mathbf{14.6\pm21.1}$ \\
4  & $26.8\pm38.4$   & $18.4\pm32.2$   & \cellcolor{green!18}$\mathbf{12.7\pm20.2}$ \\
5  & $24.75\pm35.9$  & $30.0\pm41.2$   & \cellcolor{green!18}$\mathbf{20.1\pm35.7}$ \\
6  & $16.03\pm30.6$  & $19.9\pm32.9$   & \cellcolor{green!18}$\mathbf{14.7\pm28.7}$ \\
7  & $21.41\pm33.9$  & \cellcolor{green!18}$\mathbf{13.1\pm22.7}$ & $14.4\pm35.3$ \\
8  & $23.62\pm35.0$  & $21.3\pm33.4$   & \cellcolor{green!18}$\mathbf{15.3\pm24.3}$ \\
9  & $81.18\pm247.7$ & $75.3\pm242.5$  & \cellcolor{green!18}$\mathbf{23.0\pm40.6}$ \\
10 & $20.99\pm28.7$  & \cellcolor{green!18}$\mathbf{12.0\pm23.5}$ & $20.7\pm34.2$ \\
\bottomrule
\end{tabular}
\vspace{-30pt}
\end{wraptable}

\begin{wrapfigure}{r}{0.42\linewidth}
\vspace{-25pt}
\centering
\includegraphics[width=\linewidth]{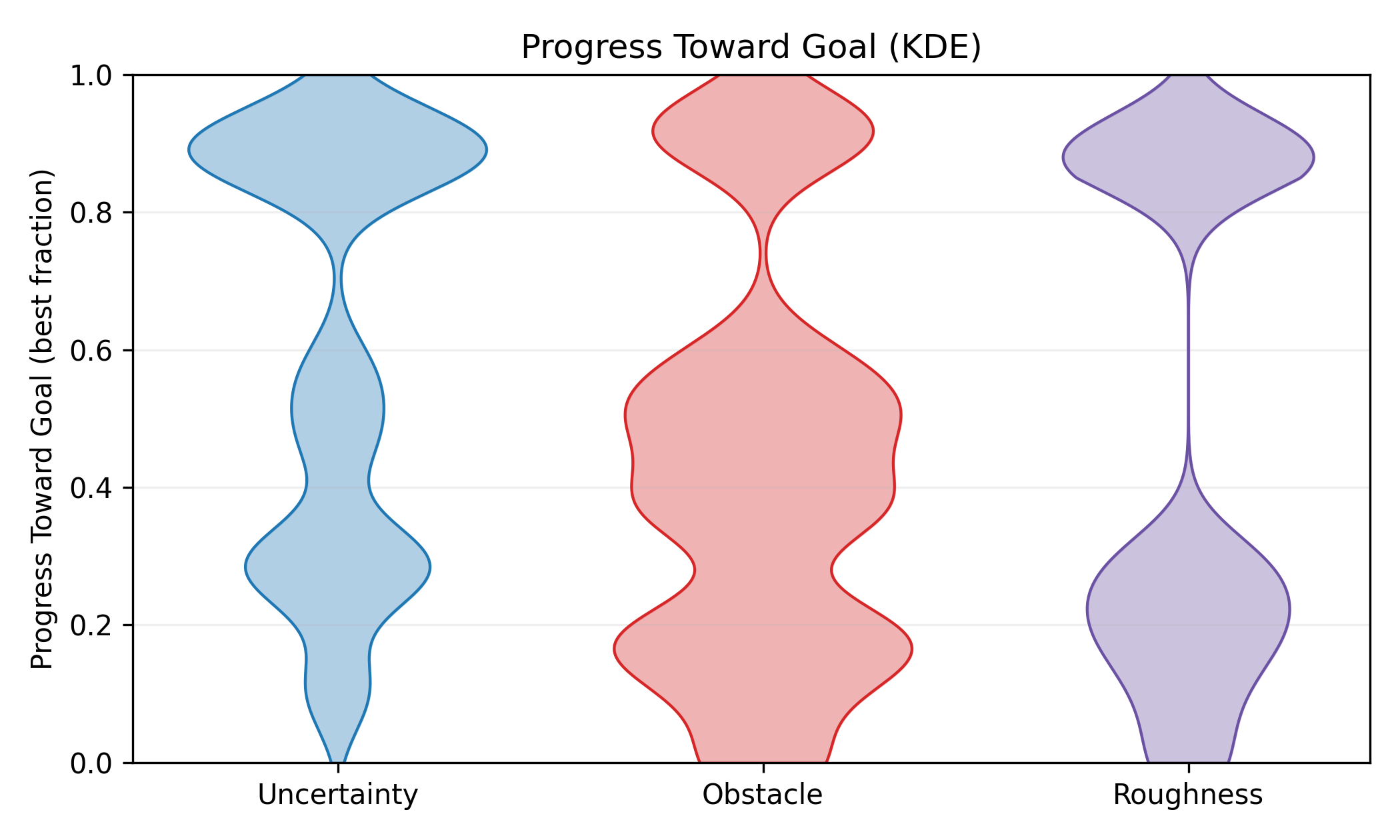}
\caption{\small KDE of goal-progress across 20 different MPPI planning runs.}
\label{fig:progress}
\vspace{-25pt}
\end{wrapfigure}

\Cref{fig:progress} shows the
distribution of goal-progress scores across 20 MPPI planning runs under each
costmap formulation. The proposed uncertainty-based costmap achieves the highest
median progress and the tightest distribution, indicating more consistent
goal-directed behavior.

The roughness baseline shows the widest spread,
reflecting its tendency to block feasible paths by treating geometrically
rough but dynamically traversable terrain as lethal. The obstacle-only
baseline achieves competitive median progress but at the cost of higher
feasibility error, as shown in~\Cref{tab:feasibility_error_summary}.

\Cref{fig:realworld_costmap} shows real-world Nav2 integration. The proposed
costmap increases cost near the small 2cm ramp placed as an anomaly within the lookahead
horizon, while the terrain-only roughness baseline fails to reflect the
motion-induced uncertainty at the same location, and the obstacle-only
Baseline does not register the ramp as a hazard.

%% file: sections/conclusion.tex
\section{Conclusion and Future Work}

We presented a feasibility-aware navigation cost for legged robots based
on task-conditioned epistemic uncertainty in future foothold prediction.
By jointly predicting footholds and their uncertainty from terrain
observations and commanded motion, the approach identifies ID and OOD
operating regimes without explicit terrain classification or retraining.
Experiments across simulation and real-world terrains show that
uncertainty-based costmaps achieve up to 37\% lower feasibility error
than geometry-only baselines and support zero-shot sim-to-real deployment
from a single trained model. The current system incorporates uncertainty only at the planning level, leaving the low-level controller unchanged and making the approach controller-agnostic and easy to integrate. Future work will use the feasibility cost for closed-loop gait adaptation and train on dynamic velocity commands to further reduce the sim-to-real gap.